# Investigating the Impact of CNN Depth on Neonatal Seizure Detection Performance

Alison O'Shea, Gordon Lightbody, Geraldine Boylan and Andriy Temko

*Abstract—* This study presents a novel, deep, fully convolutional architecture which is optimized for the task of EEG-based neonatal seizure detection. Architectures of different depths were designed and tested; varying network depth impacts convolutional receptive fields and the corresponding learned feature complexity. Two deep convolutional networks are compared with a shallow SVM-based neonatal seizure detector, which relies on the extraction of hand-crafted features. On a large clinical dataset, of over 800 hours of multichannel unedited EEG, containing 1389 seizure events, the deep 11-layer architecture significantly outperforms the shallower architectures, improving the AUC90 from 82.6% to 86.8%. Combining the end-to-end deep architecture with the feature-based shallow SVM further improves the AUC90 to 87.6%. The fusion of classifiers of different depths gives greatly improved performance and reduced variability, making the combined classifier more clinically reliable.

## I. INTRODUCTION

Neonatal seizure detection is a complex task which requires careful interpretation of the neonatal electroencephalogram (EEG) waveforms. Often seizures in newborn babies are the sole indicator of a serious neurological condition [1]. Unlike seizures in children and adults, most neonatal seizures present no physical signs and the reliable detection of seizures is only possible through EEG analysis [2]. Specially trained staff and equipment are required to detect seizures, limiting diagnosis and treatment to specialized units. These challenges have prompted research into the development of computer-based automated seizure detection algorithms. Automated seizure detection algorithms can provide objective support by alerting clinicians and aiding the early detection and treatment of seizures [3].

When detecting seizures an expert understands that background EEG is random in nature, whereas seizures represent a more ordered, rhythmic and evolving deviation from this background behaviour. This clinical knowledge has prompted the search for features which characterise the repetitiveness, order and predictability of EEG, and the application of machine learning classifiers to these features [4]. Most machine learning algorithms for neonatal seizure detection are based on hand-crafted features [5, 6] which are reliant on prior knowledge of neonatal EEG. The features are generally extracted from the time, frequency and information theory domains to provide energy, frequency, temporal and structural descriptors of neonatal EEG, giving an informative characterization of each segment of EEG.

The success of deep learning in areas such as image and audio processing [7, 8] has influenced other applications, where the feature extraction can be omitted; features can be learned directly from the data, merging feature extraction and classification in a single end-to-end optimisation routine. No domain knowledge is thus required and more importantly the resultant features extracted are not limited by this knowledge and the associated assumptions, such as stationarity, linearity, etc. One such application is EEG-based seizure detection [9-11]. Most works are motivated by the success of convolutional neural networks (CNNs) in image processing tasks where they have outperformed human benchmarks [12]. These studies follow the image processing paradigm by converting the EEG signal into a 2D representation by time-frequency or spatial-temporal analysis [13, 14].

The state-of-the-art results for the task of neonatal seizure detection have been previously obtained with a Support Vector Machine (SVM) [15]. This SVM system relies on a set of 55 hand-crafted features, which were selected and engineered specifically for the neonatal seizure detection task. The system was validated on a large clinical database of full-term neonatal EEG to confirm its robust functionality and clinically acceptable level of performance [4]. Recently, a deep CNN-based neonatal seizure detector, which works with raw EEG, has outperformed this SVM based system [16].

This work further explores the usage of deep CNNs applied to raw neonatal EEG for the task of neonatal seizure detection, by increasing the range of receptive fields and therefore the complexity of the extracted features. A novel deeper 11-layer architecture is designed and compared with a previously designed 6-layer CNN and the shallow SVM baseline. This work also explores combining deep and shallow classifiers to give more clinically robust patient independent results.

## II. METHODS AND MATERIALS

### A. SVM Baseline

The SVM system was trained and tested on a dataset of EEG collected from 18 full-term newborns, all of whom experienced seizures [15]. The 10-20 placement system, modified for neonates, was used to record 8 channels of EEG in the Neonatal Intensive Care Unit of Cork University Maternity Hospital. The data was not pre-selected and artefacts were not manually removed. The study had full

This work was supported in part by the Wellcome Trust Strategic Translational Award (098983/Z/12) and by the Science Foundation Ireland Research Centres Award (12/RC/2272). We gratefully acknowledge the support of NVIDIA Corporation with the donation of the TitanX GPU.

A. O'Shea, G. Lightbody, A. Temko, are with the Department of Electrical and Electronic Engineering, Irish Center for Fetal and Neonatal Translational Research (INFANT), University College Cork, Ireland (corresponding author's e-mail: alisonoshea@umail.ucc.ie).

G. Boylan is with the Department of Pediatrics and Child Health, INFANT Center, University College Cork, Ireland.

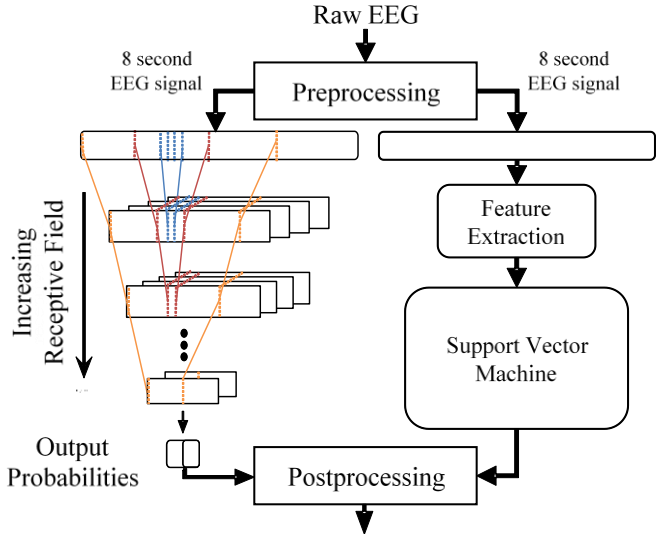

Figure 1. Comparison of shallow SVM and deep CNN algorithms. The effect of increasing the receptive field with convolutional depth can be seen. The final convolutional layer of the CNN with sample-size filters has a receptive field spanning almost the full input signal.

ethical approval from the Clinical Research Ethics Committee of the Cork Teaching Hospitals.

In the preprocessing stage, the raw EEG sampled at 256Hz is filtered between 0.5Hz and 12.8Hz, down-sampled to 32Hz and split into 8-second windows with 1 second shift. The SVM seizure detection system takes 55 features for each 8-second input, these come from the time, frequency and information theory domains.

A Gaussian kernel SVM was trained to classify each 8-second window as seizure or non-seizure. The post-processing stage includes a moving average filter of 1 minute length, adaptation to the background level of probability and a collar to compensate for the effect of smoothing. More details on the baseline system can be found in [15].

### B. Performance Assessment and Metrics

The primary performance assessment metric was the area under the receiver operating curve (AUC). The receiver operating curve is a measure of sensitivity versus specificity of a classifier.

In this paper the AUC90 metric is also reported, which measures the AUC for specificities greater than 90%. Babies are typically monitored over multiple days, thus a clinically useful system should produce a tolerably low number of false detections per hour (FD/h). The AUC90 metric reports the sensitivity for a high level of specificity, resulting in a lower number of FD/h.

In a real clinical setting, the EEG of a neonate is not available before the baby is born. The classifier thus needs to be assessed in a patient-independent way. The leave-one-patient-out cross-validation (LOO) method is the most suitable test procedure to test a patient-independent seizure detector. In this method, data from 17 newborns are used to train the classifier, and the performance is assessed on the remaining baby. This routine is repeated 18 times, so that each baby becomes the unseen test case in one experiment.

TABLE I. THE 11 LAYER CNN ARCHITECTURE. ALL CONVOLUTIONAL LAYERS ARE FOLLOWED BY RECTIFIED LINEAR UNITS. BATCH NORMALIZATION WAS USED TO SPEED UP THE TRAINING PROCESS. THE FINAL 3 LAYERS (SHADED) MAKE UP THE CLASSIFICATION PROCEDURE.

| Layer Type | Layer Parameters | Output Shape |
|---|---|---|
| Input | 256 samples | 256x1 |
| 1D Convolution | 32 1x3 filters | 254x32 |
| 1D Convolution | 32 1x3 filters | 252x32 |
| 1D Convolution | 32 1x3 filters | 250x32 |
| Batch Norm. | | |
| Average Pooling | pool 8 stride 3 | 81x32 |
| 1D Convolution | 32 1x3 filters | 79x32 |
| 1D Convolution | 32 1x3 filters | 77x32 |
| 1D Convolution | 32 1x3 filters | 75x32 |
| Batch Norm. | | |
| Average Pooling | pool 4 stride 3 | 24x32 |
| 1D Convolution | 32 1x3 filters | 22x32 |
| 1D Convolution | 32 1x3 filters | 20x32 |
| 1D Convolution | 32 1x3 filters | 18x32 |
| Batch Norm. | | |
| Average Pooling | pool 2 stride 3 | 6x32 |
| 1D Convolution | 32 1x3 filters | 4x32 |
| 1D Convolution | 2 1x3 filters | 2x2 |
| Global Average Pooling | | 2 |
| Softmax | | 2 |

### C. CNN Architectures

The CNN-based systems exploit the same pre-processing and post-processing in the SVM baseline, but the feature extraction and classification are substituted with CNNs. Single channel, preprocessed, temporal EEG signals are used as input to the CNN; omitting the feature extraction stage means that all the feature and classifier optimisation is carried out within the CNN architectures.

CNNs use local connectivity and weight sharing to extract feature hierarchies from raw data. In the designed networks small, sample-sized convolutional filters are used to give fine granularity to the features learned [17]. The receptive field of the input convolutional layer is equal to the width of a convolutional filter whereas the receptive fields of subsequent convolutional layers are hierarchically increased. Fig. 1 shows that in progressively deeper layers the complexity of features extracted increases, filters in the first layer span a narrow receptive field, but as depth increases the receptive field widens. The increased size of the receptive field allows more information to be processed by a filter bank, giving a hierarchy of feature complexity.

Previously, a 6-layer CNN was developed using sample-sized convolutional filters; each with a width of 4 samples [16]. This network outperformed the SVM baseline system, which represented the state-of-the-art in neonatal seizure detection. For this study, a CNN of 11 convolutional layers with filters of 3-sample width is developed. The details of the 11-layer architecture are disclosed in Table I. This network had slightly decreased complexity in individual filters in comparison to the 6-layer CNN, due to the smaller filter size. But the network had increased overall complexity because of the increased convolutional depth. Convolutional layers were added until the output shape of the final convolutional layer was shorter than the width of the sample sized filters i.e. no more convolutional operations were possible.

The 6-layer network has 17,058 parameters. The 11-layer network has 28,642 parameters; more parameters results in increased network capacity. The receptive field in the final convolutional layer in the 6-layer network was 47 input samples wide, whereas the largest receptive field width of the 11-layer is 212 samples. The 11-layer network can learn more simple features in the first layer (3 samples wide) and more complex features in the final layers (212 samples wide).

The decision-making output layer consists of a convolutional layer and a global average pooling layer instead of the typically used dense layers. In the final convolutional layer, the number of feature maps is equal to the number of classification categories, in this case 2 (seizure and non-seizure). After global average pooling the average values across feature maps are flattened and a softmax function is applied to the output values.

The network optimization used stochastic gradient descent with a learning rate of 0.01 and momentum of 0.9, batch size was 2048. On each iteration of the LOO routine, a randomly selected subset of balanced seizure and non-seizure data from the 17 training babies was used to train the network; the full EEG recordings from these 17 babies was used as a validation dataset. The training data contains less than 2% of the validation dataset. The network was trained for 100 epochs, after each epoch the validation AUC was calculated. The network weights which gave the best AUC on the validation data were tested on the remaining, held-out, test baby.

*D. Combination*

The CNN and SVM architectures represent diverse ways of approaching the task of neonatal seizure detection. The large contrast between domain-knowledge-led feature extraction and purely data-driven CNN feature extraction suggests the potential for both methods to complement each other. To assess the possibility of improvement through simple fusion, two blending schemes are used to combine CNN and SVM probabilities. Weighted arithmetic (1) and geometric mean (2) were applied to classifier probabilistic outputs:

$$P_{seizure} = \alpha P_{CNN} + (1 - \alpha) P_{SVM} \quad (1)$$
$$P_{seizure} = P_{CNN}^{\alpha} P_{SVM}^{1-\alpha}, \quad (2)$$

where $0 \leq \alpha \leq 1$ varies the emphasis of one classifier over the other.

The 11-layer CNN was the deepest architecture developed for this paper and it was chosen to be most suitable for combination with the SVM. The probabilistic outputs of the deep 11-layer CNN and the SVM were combined before post-processing.

III. EXPERIMENTAL RESULTS

*A. Individual Classifier Performance*

Table II shows the AUC and AUC90 results for each classifier over the dataset. The best standalone classifier is the 11-layer CNN as it achieves the highest scores in both metrics, averaged across all babies. Both the 6-layer and 11-layer CNN architectures, through end-to-end learning of discriminative patterns, outperform the SVM network on the seizure detection task. The deeper 11-layer CNN has the top AUC score in 14 babies, and it gives a relative improvement of 30% over the SVM based classifier.

TABLE II. A PER-PATIENT AND AVERAGED COMPARISON OF CLASSIFIER PERFORMANCE FOR AUC AND AUC90 METRICS.

| Baby | AUC % | | | AUC90 % | | |
|---|---|---|---|---|---|---|
| | SVM | CNN (6) | CNN (11) | SVM | CNN (6) | CNN (11) |
| 1 | 95.25 | 97.01 | **97.83** | 84.19 | 87.58 | **89.97** |
| 2 | 99.29 | 99.41 | **99.71** | 92.93 | 94.13 | **97.12** |
| 3 | 97.31 | 97.15 | **97.58** | **81.41** | 79.28 | 81.22 |
| 4 | 97.11 | 96.06 | **98.11** | 83.09 | 76.64 | **86.77** |
| 5 | 94.73 | 98.12 | **98.76** | 69.96 | 86.48 | **89.85** |
| 6 | **96.05** | 95.64 | 95.54 | **74.38** | 67.21 | 68.40 |
| 7 | 99.15 | 98.70 | **99.29** | 91.89 | 87.77 | **93.11** |
| 8 | 97.71 | 98.04 | **98.32** | 80.98 | 85.59 | **88.34** |
| 9 | 99.01 | 98.97 | **99.15** | 91.98 | **92.55** | 92.46 |
| 10 | 88.66 | 97.01 | **98.26** | 61.01 | 84.13 | **90.63** |
| 11 | **97.99** | 97.54 | 97.75 | **89.00** | 86.53 | 87.64 |
| 12 | 95.47 | 96.28 | **96.50** | 76.76 | 76.43 | **76.90** |
| 13 | 98.73 | 96.86 | **99.10** | 89.14 | 76.16 | **91.07** |
| 14 | 95.95 | 97.93 | **98.74** | 85.58 | 91.21 | **93.06** |
| 15 | 97.92 | 97.58 | **98.01** | 87.28 | 88.26 | **88.44** |
| 16 | **93.34** | 91.93 | 87.25 | 74.45 | **74.85** | 64.68 |
| 17 | 96.83 | **98.22** | 98.18 | 91.38 | **95.71** | 95.09 |
| 18 | 98.07 | 94.00 | **98.85** | 86.31 | 67.40 | **88.63** |
| Mean | 96.59 | 97.03 | **97.61** | 82.87 | 83.22 | **86.85** |
| 95%CI | 1.19 | 0.84 | 1.28 | 4.02 | 3.96 | 4.05 |

The 11-layer CNN architecture gives an improvement in AUC90 score of nearly 4% over the SVM baseline. This improvement translates to a higher epoch-based sensitivity for a set low value of FD/h; meaning more of the seizure burden will be correctly diagnosed in a clinical setting. On this dataset the SVM-baseline has a sensitivity of 70% at a rate of 0.25 FD/h [15]. By exploiting the deeper CNN architecture for the same FD/h rate the detection of correct seizure burden is increased to 81%.

Interesting examples are patient 16 and patient 10. For patient 16 the performance decreases with the increasing depth of the CNN. For this subject, the SVM system obtains the highest performance. This warrants a more thorough examination of the characteristics of this EEG to establish if the signal properties deviate from those learned by the CNNs. Conversely for patient 10, an AUC of only 88.56% is achievable by the SVM, whereas deeper CNN architectures improve the performance by almost 10%, reaching an AUC of 98.26%. These examples suggest that the two classifiers might complement each other if combined.

*B. Combined Classifier Performance*

Fig. 2 shows the performance in terms of the AUC and AUC90 of the weighted arithmetic and geometric mean as a function of α, as explained in (1) and (2). With $\alpha = 0$ and $\alpha = 1$, the performance of SVM and CNN are obtained, respectively. The bars represent the mean and the 95% confidence interval (95% CI) of the AUC across all 18 subjects.

It can be seen that an improvement in performance is achievable using a combination of the two classifiers. Both fusion techniques achieve higher scores than any of the individual classifiers. More importantly the inter-patient variation is decreased, resulting in a higher level of stability of decision making across different patients. In particular, the 95% confidence interval has decreased from 1.19% for SVM and 1.27% for CNN to as low as 0.77% for the arithmetic mean combination with α = 0.6.

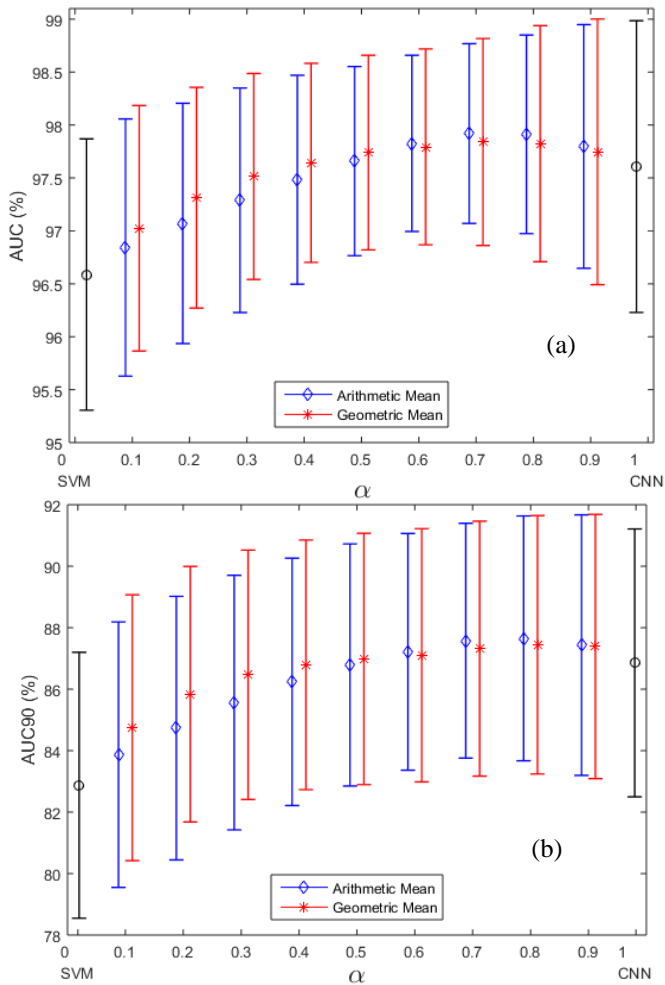

Figure 2. AUC (a) and AUC90 (b) performance of blending as a function of the combination weight, α, from (1) and (2). The leftmost (α=0) and rightmost (α=1) bars represent results of the SVM and CNN classifiers, respectively.

Both combinations of classifiers work similarly well across different thresholds, with marginally better performance achieved with the arithmetic mean. Given the higher performance of the CNN system alone, it is not unexpected that better performances are obtained for α > 0.5 which favours the CNN system. In particular, with α = 0.7 an AUC of 97.92% and an AUC90 of 87.58% are achievable. The aim of this section is to show the potential benefits of combining the end-to-end data-driven deep architecture with the prior-knowledge-based shallow SVM. In further work weights can be found and optimised on the validation data.

## IV. CONCLUSION

This work has shown that a fully convolutional network, with 11 convolutional layers running on raw EEG, can outperform a highly engineered SVM classifier, which previously obtained state-of-the-art results for the task of neonatal seizure detection. The AUC values obtained indicate that the CNN learns relevant features which have sufficient complexity to match the 55 hand-crafted features used to train the SVM system. This represents a substantial improvement over a shallower 6-layer CNN network which has a smaller range of receptive fields. These results represent the current best results for this task obtained using a single classifier.

The combination of SVM and CNN probabilistic outputs using a weighted combination has further improved the AUC and reduced inter-patient variability. The result of this work is a more clinically useful and better performing neonatal seizure detection system.